\documentclass[10pt,twocolumn,letterpaper]{article}

\usepackage[pagenumbers]{cvpr} 

\usepackage{graphicx}
\usepackage{amsmath}
\usepackage{amssymb}
\usepackage{xcolor}
\usepackage{bm}
\usepackage{multirow,multicol, makecell, booktabs}
\usepackage{wrapfig}

\usepackage[pagebackref,breaklinks,colorlinks]{hyperref}

\usepackage[capitalize]{cleveref}
\crefname{section}{Sec.}{Secs.}
\Crefname{section}{Section}{Sections}
\Crefname{table}{Table}{Tables}
\crefname{table}{Tab.}{Tabs.}

\def\cvprPaperID{9493} 
\def\confName{CVPR}
\def\confYear{2022}

\begin{document}

\title{\textsc{VL-Adapter}: Parameter-Efficient Transfer Learning \\ for Vision-and-Language Tasks}

\author{Yi-Lin Sung \qquad Jaemin Cho \qquad Mohit Bansal\\
UNC Chapel Hill\\
{\tt\small \{ylsung,jmincho,mbansal\}@cs.unc.edu}
}

\maketitle

\begin{abstract}

   Recently, fine-tuning language models pre-trained on large text corpora have provided huge improvements on vision-and-language (V\&L) tasks as well as on pure language tasks. 
   However, fine-tuning the entire parameter set of pre-trained models becomes impractical since the model size is growing rapidly. Hence, in this paper, we introduce adapter-based parameter-efficient transfer learning techniques to V\&L models such as VL-BART and VL-T5.
   We evaluate our methods in a unified multi-task setup on both image-text and video-text benchmarks. For the image-text tasks, we use four diverse V\&L datasets: VQAv2, GQA, NLVR$^{2}$, and MSCOCO image captioning. For video-text tasks, we use TVQA, How2QA, TVC, and YC2C.
   With careful training and thorough experiments, we benchmark three popular adapter-based methods (Adapter, Hyperformer, Compacter) against the standard full fine-tuning and the recently proposed prompt-tuning approach. We also enhance the efficiency and performance of adapters by sharing their weights to attain knowledge across tasks. Our results demonstrate that training the adapter with the weight-sharing technique (4.18\% of total parameters for image-text tasks and 3.39\% for video-text tasks) can match the performance of fine-tuning the entire model.
    Lastly, we present a comprehensive analysis including the combination of adapter and task-specific prompts and the impact of V\&L pre-training on adapters.\footnote{The code for our CVPR 2022 paper is available at: \url{https://github.com/ylsung/VL_adapter}.}

\end{abstract}

\section{Introduction} \label{sec:intro}

\begin{figure}
    \centering
    \includegraphics[width=\columnwidth]{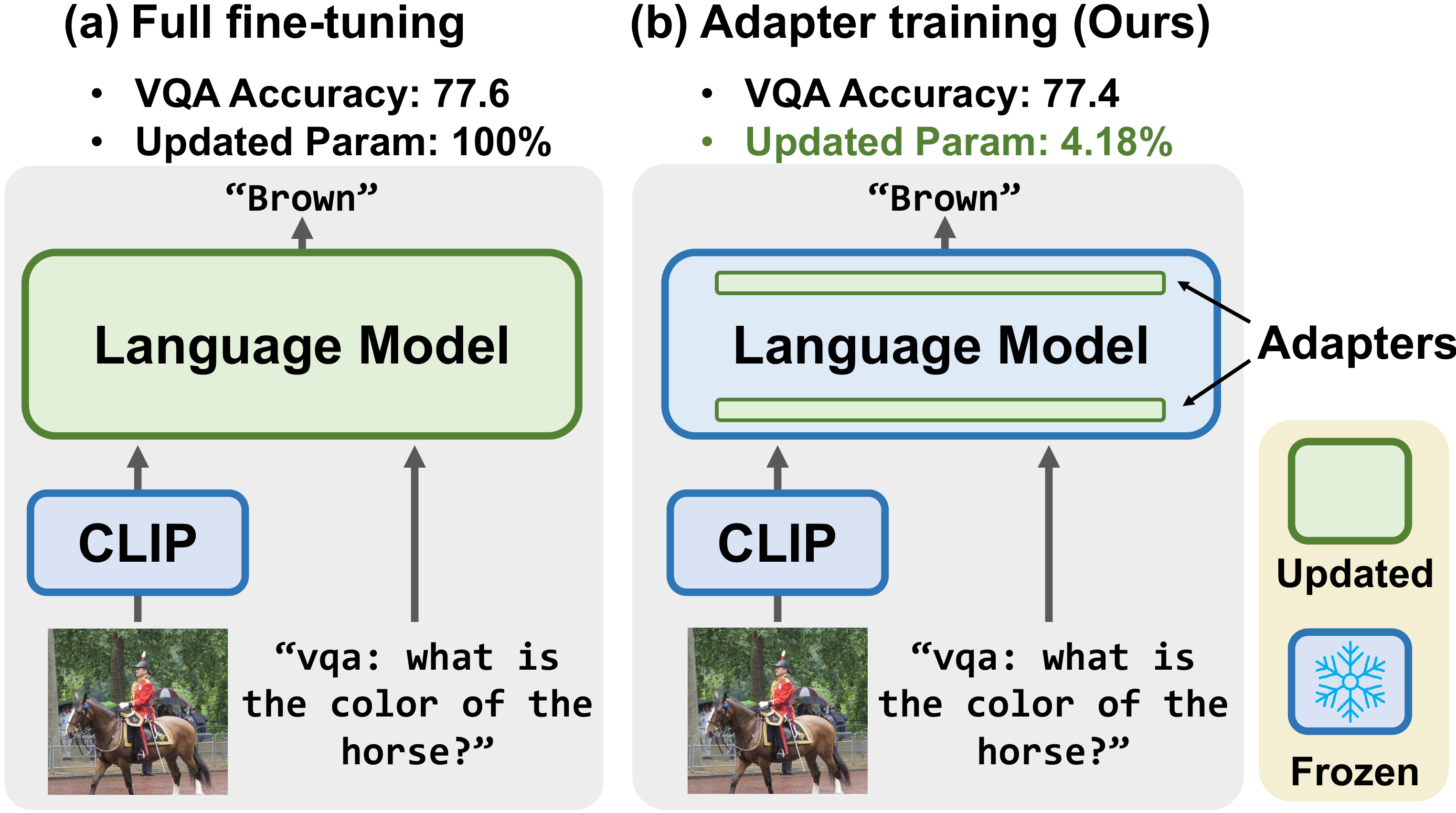}
    \caption{
    Comparison of (a) full fine-tuning and our (b) adapter training for V\&L tasks. By updating only a small set of adapter parameters, we can achieve similar performance to full fine-tuning. We experiment with our adapter training on diverse image-text and video-text benchmarks, and here we illustrate the VQA task as an example V\&L task.
    }
    \label{fig:teaser}
\end{figure}

\begin{figure*}
    \centering
    \includegraphics[width=0.85\linewidth]{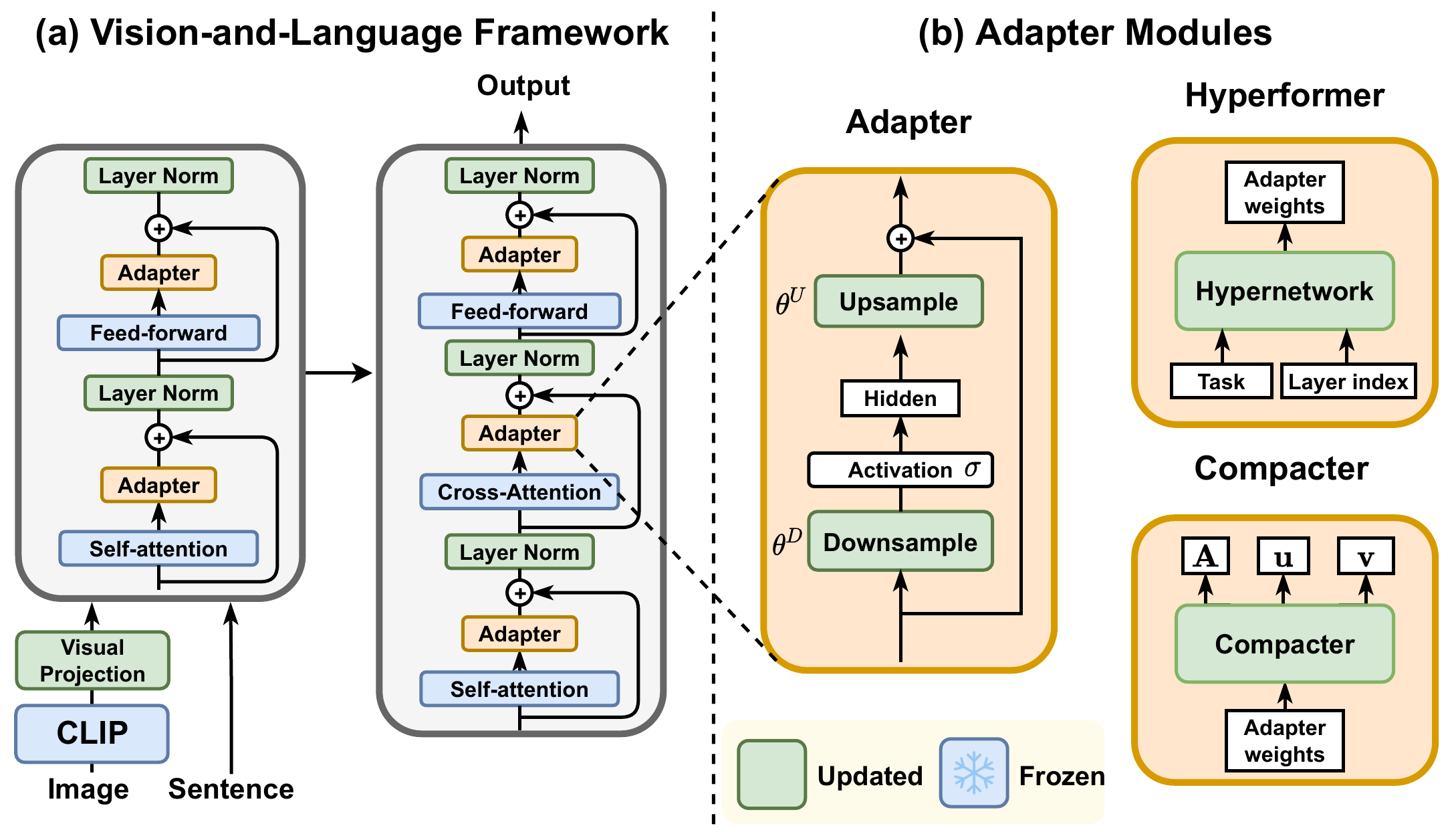}
    \vspace{-4pt}
    \caption{Illustration of (a) our unified framework for V\&L tasks (Sec.~\ref{ssec:framework}) and and (b) three-adapter based modules (Sec.~\ref{ssec:adapters}).
    Green color refers to trainable parameters and blue color refers to frozen ones.
    }
    \label{fig:architecture}
    \vspace{-5pt}
\end{figure*}

Following the success in the language domain \cite{Devlin2019BERTPO,DBLP:journals/corr/abs-1907-11692,DBLP:journals/corr/abs-1910-13461,DBLP:journals/corr/abs-1910-10683,Radford2019LanguageMA,DBLP:journals/corr/abs-2005-14165}, large-scale pre-training of vision-and-language (V\&L) models has become a standard framework to tackle V\&L tasks \cite{Lu2019ViLBERTPT,Tan2019LXMERTLC,DBLP:journals/corr/abs-2004-00849,DBLP:journals/corr/abs-1909-11740,DBLP:journals/corr/abs-2102-02779,DBLP:journals/corr/abs-2107-06383}.
In such frameworks, V\&L models, which are usually the combination of vision encoders and language models, are first pre-trained on large-scale unlabeled data, then fine-tuned for downstream V\&L tasks.
This is the standard strategy to fuse the knowledge of vision-and-language to the language model.
However, given that such models' size grows very rapidly nowadays, either pre-training or fine-tuning of the V\&L model can still contribute to an unignorable, large memory and storage cost.
For instance, if we use GPT-3 \cite{DBLP:journals/corr/abs-2005-14165} with 175B parameters as a backbone of V\&L model, we would need 700 GB of memory to store its entire parameters.\footnote{$(175 \times 10^{9}) \times 4 \text{(bytes)} \times \frac{1}{10^{-9}} \text{(GB/bytes)} = 700 \text{(GB)}$}
To address this problem, recently, 
several parameter-efficient training methods have been proposed \cite{houlsby2019parameter,DBLP:journals/corr/abs-2106-04489,DBLP:journals/corr/abs-2106-04647,DBLP:journals/corr/abs-2101-00190,DBLP:journals/corr/abs-2106-10199,DBLP:journals/corr/abs-2012-07463,DBLP:journals/corr/abs-2106-09685,sung2021training}. Among them, adapter \cite{houlsby2019parameter} and its variants \cite{DBLP:journals/corr/abs-2106-04489,DBLP:journals/corr/abs-2106-04647} are widely used in the NLP domain and applied to different architectures. Adapter is a small module added to intermediate layers of the model (which is illustrated in \Cref{fig:architecture}), which allows to achieve as high performance as full fine-tuning (i.e., updating all parameters), by fine-tuning only a small set of parameters. 
Moreover, this also shows that it is possible to use a few parameters to learn the information fusion of vision and language without losing performance.
Despite adapters having achieved success in text classification \cite{houlsby2019parameter,DBLP:journals/corr/abs-2106-04489,DBLP:journals/corr/abs-2106-04647} and image-text alignment \cite{anonymous2022how}, to the best of our knowledge, no work has utilized this efficient method for more challenging downstream V\&L problems, such as visual/video question answering and image/video captioning. Besides, V\&L models often come with expensive computations by combining the knowledge of two input modalities. Hence, we investigate the application of adapter-based parameter-efficient training techniques to V\&L tasks.

We aim to efficiently tune language models on diverse downstream V\&L tasks while achieving performance comparable to full fine-tuning.
For this, we analyze these parameter-efficient training techniques in a unified multi-task learning setup, and we benchmark different adapter \cite{houlsby2019parameter,DBLP:journals/corr/abs-2106-04489,DBLP:journals/corr/abs-2106-04647} and prompt-based methods \cite{Lester2021ThePO}. For our V\&L model, following Cho \etal{}~\cite{DBLP:journals/corr/abs-2102-02779}, we adopt encoder-decoder language models (BART \cite{DBLP:journals/corr/abs-1910-13461} and T5 \cite{DBLP:journals/corr/abs-1910-10683}) that tackle multiple V\&L tasks as text generation to avoid designing task-specific architectures. We use CLIP \cite{DBLP:journals/corr/abs-2103-00020}, a pretrained image-text alignment model, as our visual encoder for the ease of doing V\&L pre-training. To inform the model about which task it is going to perform, we follow \cite{DBLP:journals/corr/abs-1910-10683, DBLP:journals/corr/abs-2102-02779} to add task-specific (text) prompts to the front of the input sentence (e.g., ``vqa: [Q]'' for VQA). We then insert Adapter~\cite{houlsby2019parameter} and its variants, Hyperformer \cite{DBLP:journals/corr/abs-2106-04489} and Compacter \cite{DBLP:journals/corr/abs-2106-04647}, into the model to perform parameter-efficient training.
Hyperformer and Compacter are recently proposed \textit{state-of-the-art} approaches: Hyperformer improves the efficiency of adapters by generating their weights via a hyper-network, while Compacter reduces the parameters by utilizing Kronecker products and low-rank parameterization for the adapters' weights. We also compare adapter-based approaches with prompt tuning \cite{Lester2021ThePO}, which adds trainable prompts to the input. We show the high-level concept of our work in \Cref{fig:teaser}. Practically, adapter training involves parameter updates of adapter modules, layer normalization layers, and the visual projection layer (see \Cref{ssec:framework} and \Cref{fig:architecture} for more details). Since we tackle multiple tasks simultaneously \cite{Pilault2021ConditionallyAM,DBLP:journals/corr/abs-2106-04489}, we also explore taking advantage of the sharing of information between tasks on adapters and prompts.
Specifically, we make some of the trainable parameters to be shareable to learn cross-task information while reserving the rest of them for task-specific information. With this technique, the number of trainable parameters can be reduced even further.

We conduct our experiments and analysis on four diverse image-text tasks: VQAv2 \cite{Goyal2017MakingTV}, GQA \cite{DBLP:journals/corr/abs-1902-09506}, NLVR$^{2}$ \cite{DBLP:journals/corr/abs-1811-00491}, and MSCOCO captioning \cite{DBLP:journals/corr/ChenFLVGDZ15}. For completeness, we also verify the effectiveness of our framework on four video-text tasks: TVQA \cite{Lei2018TVQALC}, How2QA \cite{Li2020HeroHE}, TVC\cite{Lei2020TVRAL}, and YC2C \cite{Zhou2018TowardsAL}.
Overall, the performance of the three adapter-based approaches closes the gap between which of full fine-tuning.
In our experiments, Compacter does not stand out in terms of efficiency since we remove the low-rank approximation for trading performance. Hyperformer is more efficient than adapters, but we eventually show \textit{our adapter training with the weight-sharing technique can achieve the same performance as full fine-tuning while only updating 4.18\% of the entire parameters for image-text tasks (and 3.39\% for video-text tasks)}.
Next, we compare the fine-tuning and freezing of the CLIP parameters, where the latter achieves a better trade-off between performance and parameter efficiency. We also present a detailed analysis to understand the contribution of each trainable component in adapters, as well as the different parameter-sharing mechanisms. We find that using a single set of adapter modules across all tasks achieves the best results and accuracy-efficiency trade-off, showing the possibility of pursuing efficiency with simplicity (Fig.~\ref{fig:adapters_comparison}). Since most of the experiments are based on the pre-trained weights accompanied with the models (e.g., CLIP pre-trained weights for ResNet and BART pre-trained weights), we also demonstrate that the results of training adapters on top of V\&L pre-trained weights can match or even exceed which of the full fine-tuning counterpart. While we conduct most of our experiments with the V\&L generation framework, we also extend the adapter training to CLIP-ViL \cite{DBLP:journals/corr/abs-2107-06383}, which is one of the SOTA discriminative V\&L approaches.
Lastly, we report the results of comprehensive hyperparameter search in \Cref{sec:hp_search},
hoping that they will be useful for related research on parameter-efficient training.

Our contributions could be summarized as:
(1) the first work benchmarking different types of parameter-efficient training techniques (Adapter, Hyperformer and Compacter) for diverse challenging downstream image-text and video-text tasks;
(2) empirical demonstration of adapters reaching the performance of full fine-tuning while updating only 3.39-4.18\% of the parameters;
(3) comprehensive analysis on the design of freezing CLIP, impact of different architectural components, weight-sharing techniques, task-specific prompts, and vision-language pretraining.

\section{Related Work}
\label{sec:related works}

Our research is built upon previous works about language models, V\&L models, and parameter-efficient training. In this section, we introduce previous literature and discuss their similarities and differences with this work.

\vspace{3pt}
\noindent\textbf{Language Model Pre-training.}
The workflow of pre-training and fine-tuning has become a popular paradigm for solving many downstream tasks in the NLP field. Several papers accordingly propose new architectures and objectives to model language, such as ELMo \cite{Peters2018DeepCW}, BERT \cite{Devlin2019BERTPO}, RoBERTa \cite{Liu2019Roberta}, GPT series \cite{Radford2018ImprovingLU,DBLP:journals/corr/abs-2005-14165, Radford2019LanguageMA}, and encoder-decoder versions such as BART \cite{DBLP:journals/corr/abs-1910-13461} and T5 \cite{DBLP:journals/corr/abs-1910-10683}. Comprehensive studies \cite{DBLP:journals/corr/abs-1910-10683} have shown the effectiveness of the encoder-decoder model compared to other architectures. Hence, we choose BART and T5 as our text generative model.

\vspace{3pt}
\noindent\textbf{V\&L Pre-training.}
To tackle V\&L tasks, most existing approaches combine respective models specialized for either pure language or pure vision tasks as a V\&L model. ViLBERT \cite{Lu2019ViLBERTPT}, UNITER \cite{DBLP:journals/corr/abs-1909-11740} and LXMERT \cite{Tan2019LXMERTLC} use the faster R-CNN \cite{Ren2015FasterRT} object detector to extract the bottom-up \cite{Anderson2018BottomUpAT} features from the image and provide them with text features to a cross-modality transformer to solve visual question answering and image-text alignment tasks.
Cho \etal \cite{DBLP:journals/corr/abs-2102-02779} adopt encoder-decoder language models \cite{DBLP:journals/corr/abs-1910-13461,DBLP:journals/corr/abs-1910-10683} to generalize V\&L tasks to text generation with a task-agnostic architecture, which combines faster R-CNN object detector and T5 (or BART).
In practice, the R-CNN is usually frozen in most of the V\&L models because the end-to-end training of the R-CNN and language models is unstable.
This prevents R-CNN from adapting its weights to V\&L tasks.
PixelBERT \cite{DBLP:journals/corr/abs-2004-00849} replaces R-CNN with plain ResNet \cite{he2016deep}, and demonstrates the advantage of including the vision model in training.
Inspired by the success of pre-training with web-scale unlabeled data in the NLP field, Radford \etal \cite{DBLP:journals/corr/abs-2103-00020} propose CLIP to show the success can be transferred to the V\&L field. With the pre-training on 400 million image-text pairs, CLIP has a rich cross-modal representation and can solve a wide range of tasks without additional supervision. CLIP-ViL \cite{DBLP:journals/corr/abs-2107-06383} explores the advantage of using a CLIP visual encoder (ResNet or ViT \cite{DBLP:journals/corr/abs-2010-11929}) for V\&L tasks.

Although language models are tuned for downstream tasks in previous works, some recent research \cite{DBLP:journals/corr/abs-2106-13884,Yang2021AnES} attempts to freeze large language models (e.g., GPT-3) to achieve zero-shot learning for V\&L tasks. This line of research focuses on how to map images to the inputs that the language model can use. Frozen \cite{DBLP:journals/corr/abs-2106-13884} achieves this by jointly training an NF-ResNet-50 \cite{DBLP:journals/corr/abs-2102-06171} and frozen GPT-3 with the Conceptual Captioning dataset \cite{sharma-etal-2018-conceptual}. Instead of aligning images to text features, Yang \etal \cite{Yang2021AnES} directly utilize a pre-trained image captioner to transform images to text, which is a useful resource for a language model, and they demonstrate the effectiveness of this framework on visual question answering tasks. 
Our V\&L model is the combination of CLIP and BART (or T5). We conduct a thorough ablation study to test the performance of four different training scenarios: all possible pairs of training or freezing CLIP and BART. The results in \Cref{ssec:freezing resnet} show that fine-tuning (i.e., not freezing) the language model is crucial to achieve competitive performance on diverse downstream V\&L tasks, which is why we focus on how to achieve this much more efficiently via different adapter methods and knowledge sharing across tasks. Our results also show that training the BART model only has the best trade-off between performance and parameter efficiency, hence, even though the architecture is end-to-end trainable, we decide to freeze the CLIP to fulfill our goal of parameter-efficient training.

\vspace{3pt}
\noindent\textbf{Parameter-Efficient Training.}
As the size of recent models increases rapidly, updating the models in parameter-efficient ways becomes crucial.
Recently, three types of methods have been proposed:
(1) only updating newly added parameters (added either to the input or model); \cite{Lester2021ThePO,houlsby2019parameter,DBLP:journals/corr/abs-2106-04489,DBLP:journals/corr/abs-2106-04647,DBLP:journals/corr/abs-2101-00190};
(2) sparsely updating a small number of parameters of the model; and \cite{sung2021training,DBLP:journals/corr/abs-2106-10199,DBLP:journals/corr/abs-2012-07463}
(3) low-rank factorization for the weights to be updated \cite{DBLP:journals/corr/abs-2106-09685}.
\cite{he2021unified,mao2021unipelt} combine such approaches to propose a unified parameter-efficient training framework.
Among these approaches, adapters, which belong to the first category, are widely used in computer vision \cite{Rebuffi2018EfficientPO, Rebuffi2017LearningMV} and natural language processing \cite{houlsby2019parameter,DBLP:journals/corr/abs-2106-04489,DBLP:journals/corr/abs-2106-04647}.
While adapters add additional parameters into models, prompt-based approaches instead add trainable parameters into the inputs \cite{Lester2021ThePO, DBLP:journals/corr/abs-2101-00190,Gu2021PPTPP}, and experiments have shown their value in language tasks.
Some concurrent works also extend parameter-efficient techniques to CLIP models \cite{Zhou2021LearningTP,anonymous2022how,Zhang2021TipAdapterTC}.
However, they mainly tackle image-text alignment problems, while in this paper we aim at more challenging downstream V\&L tasks, such as visual/video question answering, visual reasoning, and image/video captioning.
The typical usage of adapters is training them independently on different tasks. This training manner prevents these adapters from learning the shared information across tasks. In this paper, we find that making these convenient plug-and-play adapters shareable improves the performance of a low resource dataset and reduces the overall trainable parameters.

\section{Methods} \label{sec:method}

Our main contribution is to exploit adapter-based methods to efficiently fine-tune generative models in the multi-task setting. We explore several \textit{state-of-the-art} variations of adapters (Adapter, Hyperformer, Compacter) on various V\&L tasks and show that vanilla adapters are the best among them. We further demonstrate that sharing adapters across tasks can boost performance to match full fine-tuning results and further improve the parameter efficiency.

\subsection{Unified Framework for V\&L Tasks} \label{ssec:framework}

We illustrate our V\&L model in \Cref{fig:architecture}(a).
We follow \cite{DBLP:journals/corr/abs-2102-02779} to unify V\&L tasks to a text generation problem. Our V\&L model is a combination of CLIP and BART (T5), and therefore we name our 
base architecture CLIP-BART (CLIP-T5). To be more specific, assuming that we have a pair of an image (or video) $\bm{x}^I$ and a sentence $\bm{x}^S$ (e.g. the question in VQA) as the input for our model, our goal is to maximize the agreement between the model's output and the text label of $M$ tokens $\bm{y} = (y_1, y_2, ..., y_M)$ (e.g., the ground truth answer in VQA).
Regarding architectures, we use an encoder-decoder language model (parameterized by $\theta^{L}$) as our main generative model.
We connect a CLIP (parameterized by $\theta^{V}$) and a visual projection layer (parameterized by $\theta^{V \rightarrow L}$; it projects the visual representation to the correct dimension for language model) to the model for extracting the visual representation from input images and feed the concatenation of visual representation and sentence representation to the encoder-decoder model. 
The multi-head attention layers inside the encoder learn the cross-modality representations and the decoder utilizes them to generate the targeted text by maximizing its likelihood.
Note that the sentence representation is the output of an embedding layer, and positional embeddings are added to both visual and sentence representation. For simplicity, we omit the notations for the embedding layer and positional embedding since they can be viewed as part of $\theta^{L}$ and $\theta^{V}$. With the datum and model's parameters, our goal is to minimize the cross entropy (CE) loss:

\begin{align}
    \label{eq:loss function}
    \begin{split}
    & l \left(\bm{x}^I, \bm{x}^S, \bm{y}; \theta^{L}, \theta^{V}, \theta^{V \rightarrow L}\right) \\
    & = \text{CE}( f_{\theta^{L}}(\bm{x}^{V \rightarrow L}, \bm{x}^S), \bm{y}) \\
    & = - \sum_{i=1}^{M} y_i \log( f_{\theta^{L}}(\bm{x}^{V \rightarrow L}, \bm{x}^S)_i)
    \end{split}
\end{align}
where $f_{\theta}$ denotes a function parameterized by $\theta$, and $\bm{x}^{V \rightarrow L}$ is the projected visual representation, that is, $\bm{x}^{V \rightarrow L} = f_{\theta^{V \rightarrow L}} (f_{\theta^{V}}(\bm{x}^I))$.

We next introduce our multi-task setup for the unified generative model. Benefiting from the unified format, we can construct a universal dataset $\mathcal{D}$ from $N$ given V\&L datasets, $\mathcal{D}_1$, $\mathcal{D}_2$, ..., $\mathcal{D}_N$. Thus, we optimize our parameters by minimizing the averaging loss of datum in $\mathcal{D}$:
\begin{align}
    \begin{split}
    & \mathcal{L} \left(\mathcal{D}; \theta^{L}, \theta^{V}, \theta^{V \rightarrow L}\right) \\
    & = \frac{1}{\left| \mathcal{D} \right|} \sum_{(\bm{x}^I, \bm{x}^S, \bm{y}) \in \mathcal{D}} l \left(\bm{x}^I, \bm{x}^S, \bm{y}; \theta^{L}, \theta^{V}, \theta^{V \rightarrow L}\right)
    \end{split}
\end{align}
Our trainable parameters are the union of $\theta^{L}$, $\theta^{V}$, and $\theta^{V \rightarrow L}$. As deep learning models have grown rapidly in recent times, updating and storing the whole parameters of either visual or language models can be inefficient. Therefore, this motivates us to introduce adapter-based approaches into V\&L models for parameter-efficient tuning.

\subsection{Adapters for V\&L Models} \label{ssec:adapters}

\vspace{2pt}
\noindent\textbf{Adapters.} \Cref{fig:architecture}(b) left. Adapters \cite{houlsby2019parameter} are sub-networks with small parameters that are inserted after every attention and feed-forward layer in a model.
With adapters, the models learn downstream tasks by updating only a small number of parameters.
The adapters consist of a pair of downsampling and upsampling layers, and a residual connection.
To be more specific, we denote the input of the adapter as $\bm{x} \in \mathbb{R}^{d_i}$, and the weight matrices for downsampling and upsampling layers to be $\theta^{D} \in \mathbb{R}^{d_i \times d}$ and $\theta^{U} \in \mathbb{R}^{d \times d_i}$, where $d_i$ and $d$ are the input and hidden dimensions, respectively. The mechanism of adapters is defined as:
\begin{equation}
    \label{eq:adapter}
    h = f_{\theta^U}(\sigma( f_{\theta^D}(\bm{x}))) + \bm{x}
\end{equation}

where $\sigma(\cdot)$ is an activation function, and we use GELU \cite{DBLP:journals/corr/HendrycksG16} in this paper. With adapters, the parameter complexity (i.e., the number of added parameters) is $\mathcal{O}(d_i d)$, and it usually is $2 \sim 3\%$ of the whole model's parameters. Note that all layer normalization layers are also updated to adapt to the data distribution of downstream data.

\vspace{3pt}
\noindent\textbf{Hyperformers.} \Cref{fig:architecture}(b) top-right. The typical usage of adapters is to separately train one adapter for one task. In that fashion, adapter modules are independent across tasks, preventing the possibility of reducing the required parameters by sharing the weights for similar tasks. Hence, in order to make the adapter module even more efficient, we extend the Hyperformer \cite{DBLP:journals/corr/abs-2106-04489} to a V\&L architecture. More specifically, we maintain a hyper-network that is shared over tasks to generate the adapters' weights conditioned on the task and the index of the layer. Suppose that we have $N_T$ tasks at hand, the number of layers of the model is $N_L$, and we can use $\bm{t}_1, \bm{t}_2, ..., \bm{t}_{N_T} \in \mathbb{R}^{d_e}$ and $\bm{l}_1, \bm{l}_2, ..., \bm{l}_{N_L} \in \mathbb{R}^{d_e}$ to represent their $d_e$ dimensional embeddings. Hyperformer is composed of a two-layer task projector network $\theta^{T} \in \mathbb{R}^{d_e \times 2 \times d_p}$ and the hyper-network $\theta^{H} \in \mathbb{R}^{d_p \times (2 \times d \times d_i)}$, which aims to generate the weights for the upsampling and downsampling layer in adapters based on the projected embedding. Without loss of generality, to generate the adapter's weights in the $i^{th}$ layer for $j^{th}$ task, the generation process is:
\begin{equation}
    [\theta^{D}, \theta^{U}] = f_{\theta^{H}}(f_{\theta^{T}}([\bm{t}_j, \bm{l}_i]))
\end{equation}
where $[\cdot]$ refers to concatenation.
Note that to save memory with Hyperformer, the number of trainable parameters of Hyperformer needs to be smaller than that of adapters, namely, $|\theta^{H}| + |\theta^{T}| + N_T|\bm{t}| + N_L|\bm{l}| < N_T N_L (|\theta^{U}| + |\theta^{D}|)$. In general, we have $N_T|\bm{t}|, N_L|\bm{l}|,  |\theta^{T}| \lll |\theta^{H}|$, so we can further induce the appropriate range of $d_p$, which is $d_p < N_T N_L$.

\vspace{3pt}
\noindent\textbf{Compacters.} \Cref{fig:architecture}(b) bottom-right. Although adapters have attained great success on parameter-efficient training, they still have redundant parameters and usually underperform full fine-tuning. Compacter hence is introduced by \cite{DBLP:journals/corr/abs-2106-04647} 
to solve the issues with the matrix decomposition and parameter sharing, and eventually, Compacter has been shown to have a better trade-off between performance and efficiency compared to adapters. In the following, we demonstrate the mechanism of Compacter with the weights of the downsampling layer. First, Compacter introduces \textit{parameterized hypercomplex multiplication layers} (PHM layers) \cite{Zhang2021BeyondFL}, whose parameters are the decomposition of $\theta^D \in \mathbb{R}^{d_i \times d}$ to the sum of $k$ Kronecker products:
\begin{equation}
    \theta^{D} = \sum_{i=1}^{k} A_i \otimes B_i
\end{equation}
where $A_i \in \mathbb{R}^{k \times k}$, $B_i \in \mathbb{R}^{\frac{d_i}{k} \times \frac{d}{k}}$. The parameter complexity of the PHM layer is $\mathcal{O}(\frac{d_id}{k})$, reducing the cost by at most $\frac{1}{k}$.
To further improve the efficiency of PHM layers, Compacter shares the parameters of a smaller matrix $A_i$ across all layers, and decomposes the bigger matrix $B_i$ even more with low-rank parameterization.
Specifically, the matrix $B_i$ is approximated to be a low-rank matrix, which is the product of two low-rank matrices, $\bm{u}_i \in \mathbb{R}^{\frac{d_i}{k} \times r}$ and $\bm{v}_i \in \mathbb{R}^{r \times \frac{d}{k}}$, where $r$ is the matrix's rank. This results in \textit{low-rank parameterized hypercomplex multiplication layer} (LPHM):
\begin{equation}
    \theta^{D} = \sum_{i=1}^{k} A_i \otimes B_i = \sum_{i=1}^{k} A_i \otimes (\bm{u}_i\bm{v}_i)
\end{equation}
Empirically, $r=1$ is sufficient to achieve competitive performance, obtaining the complexity of the LPHM layer as $\mathcal{O}(\frac{d+d_i}{k})$.
Nevertheless, we find sharing matrix and low-rank decomposition in LPHM layers both severely hurt the performance of V\&L tasks, so we remove them and use the PHM layers instead.

\begin{figure}
    \centering
    \includegraphics[width=0.87\linewidth]{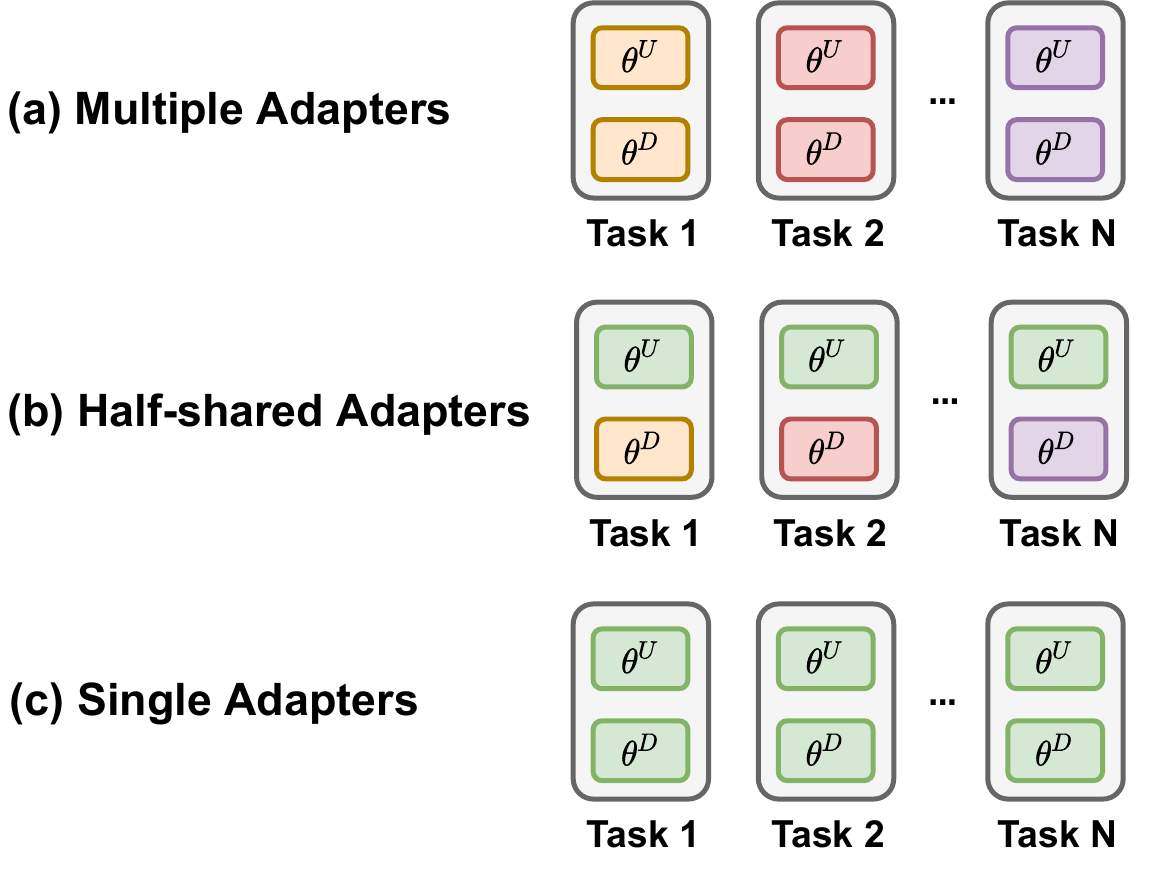}
    \vspace{-5pt}
    \caption{Illustration of different weight sharing methods for adapters. Parameters with same color are shared across tasks.
    }
    \label{fig:share}
    \vspace{-5pt}
\end{figure}

\vspace{3pt}
\noindent\textbf{Shared-Weight Adapters.} \Cref{fig:share}.
Inspired by Hyperformer, we explore the sharing of information between $N_T$ tasks in vanilla adapters with the weight-sharing technique. We denote $\Theta = \{\Theta^D, \Theta^U\}$, is the collection of all inserted adapter modules' weights in the model, and $\Theta^D$ ($\Theta^U$) is the subset that only includes downsampling (upsampling) layers in adapters. As we have mentioned earlier, adapters are trained independently, so we have unique $\{\Theta^{D}_i, \Theta^{U}_i\}$ for the $i^{th}$ task. To enable the adapter to learn cross-task information, we make part of the weights of the adapter to be shareable. For instance, we can make $\Theta^{D}_i$ equal to $\Theta^{D}_j$ ($i \neq j$), and the rest of parameters ($\Theta^{U}_i$) can still learn the task-specific information for the $i^{th}$ task. Note that we also consider the extreme case of using a single adapter for all tasks ($\Theta_i = \Theta_j$). We illustrate different weight sharing methods for adapters in \Cref{fig:share}.

\vspace{3pt}
\noindent\textbf{Where to Add Adapters?} 
Our goal is to apply adapters into V\&L models to efficiently update their parameters. Since our architecture is composed of both visual and language components, it is expected that adapter layers are injected into both of them.
However, we observe that even fully fine-tuning the whole model does not bring much improvement compared to only updating the language one (see details in \Cref{ssec:freezing resnet}).
Since freezing CLIP has the better trade-off between performance and parameter efficiency over training it, we eventually do not add adapter layers into CLIP.

\vspace{3pt}
\noindent\textbf{Multi-task Adapter Variants.}
We consider several approaches to using adapters. Since we are in a multi-task setting, the first two straightforward methods are training adapters and Compacter per task, and we call \textbf{Multiple Adapters} and \textbf{Multiple Compacters} (illustrated in \Cref{fig:share}(a)). To allow adapters to learn information across tasks, we use weight sharing techniques mentioned in \Cref{ssec:adapters} to form \textbf{Half-shared Adapters}, illustrated in \Cref{fig:share}(b). Notably, we can form the two types of Half-shared Adapters by sharing either upsampling layers or downsampling layers. However, we use Half-shared Adapters to represent adapters with sharing upsampling layers, since they have almost no differences in terms of accuracy and efficiency. We also consider the extreme case of training multiple tasks with one set of adapter layers, and this brings \textbf{Single Adapter} and \textbf{Single Compacter} (see \Cref{fig:share}(c) for details). Lastly, we have \textbf{Hyperformer} which essentially shares information from multiple tasks. 
Note that we always update $\theta^{V \rightarrow L}$ ($1.14\%$ of parameters) and all layer normalizations ($0.04\%$ of parameters) in the language model. We freeze the output layer, whose weights are tied with word embeddings, because it occupies about $30\%$ of the language model's parameters, and updating it doesn't come with a performance boost.

\begin{table*}[t]
    \centering
    \resizebox{0.83\textwidth}{!}{
    \begin{tabular}{l c c c c c c}
    \toprule
    \makecell{Method} &
    \makecell{Updated \\ Params \\ (\%)} & \makecell{VQA \\ Karpathy test \\ Acc. (\%)} & \makecell{GQA \\ test-dev \\ Acc. (\%)} & \makecell{NLVR$^{2}$ \\ test-P \\ Acc. (\%)} & \makecell{COCO Cap. \\ Karpathy test \\ CIDEr} & \makecell{Avg.} \\
    \midrule
    VL-BART \cite{DBLP:journals/corr/abs-2102-02779}            &  100.00 & \textbf{67.8} & \textbf{57.3} & 72.3 & 109.4 & 76.7 \\
    CLIP-BART \\
    + Full fine-tuning & 100.00 & 67.6 & 56.7 & \textbf{73.0} & \textbf{112.9} & \textbf{77.6} \\
    + Multiple Adapters   & 12.22 & 65.4 & 54.0 & 69.8 & 114.3	& 75.9 \\
    + Half-shared Adapters   & 8.36 & 65.2 & 53.4 & 71.2 & 113.7 & 75.9  \\
    + Single Adapter      & 4.18 & \textbf{65.9} & \textbf{54.5} & \textbf{74.2} & 114.9 & \textbf{77.4} \\
    + Hyperformer         &  5.79 & 65.1 & 53.4 & 72.3 & 114.6 & 76.4 \\
    + Multiple Compacters & 7.05 & 64.6 & 53.4 & 69.1	& \textbf{116.0} & 75.8 \\
    + Single Compacter   & 2.70 & 64.2 & 53.3 & 	71.7 & 114.1 & 75.8 \\
    + Multiple LoRA & 17.72 & 65.5 & 53.0 & 62.8 & 115.4 & 74.2 \\
    + Single LoRA & 5.93 & 65.2 & 53.6 & 71.9 & 115.3 & 76.5 \\
    + Multiple Prompts   & 4.53 & 43.8 & 38.1 & 51.1 & 104.6 & 59.4 \\
    + Single Prompt   & 2.00 & 44.0 & 36.3 & 51.8 & 103.9 & 59.0 \\
    \bottomrule
    \end{tabular}
    }
    \caption{
    The multi-task evaluation results on VQA, GQA, NLVR$^{2}$ and COCO Caption between full fine-tuning, adapter-based approaches, prompt-tuning, LoRA, and VL-BART. We bold the highest scores separately for approaches that are with or without parameter-efficient training techniques. We also report the results of the test-dev split on VQA in \Cref{sec:test-dev vqa}, and the trend is similar to using the Karpathy test set. Note that we don't use V\&L pre-training for every model.
    }
    \label{tab:main table}
    \vspace{-5pt}
\end{table*}

\begin{table}[t]
    \centering
    \resizebox{0.8\linewidth}{!}{
    \begin{tabular}{l c c c}
    \toprule
    \makecell{Method} & \makecell{Updated \\ Params (\%)} &
    \makecell{VQA \\ test-std} & \makecell{GQA \\ test-std} \\
    \midrule
    CLIP-BART \\
    + Full fine-tuning & 100.00 & 70.1 & 52.5\\
    + Single Adapter & 4.18 & 68.3 & 50.9 \\
    + Single LoRA & 5.93 & 67.3 & 50.0 \\
    + Single Prompt & 2.00 & 45.3 & 37.3 \\
    \bottomrule
    \end{tabular}
    }
    \caption{
    Leaderboard results of test-std split for representative approaches from different method families.
    }
    \label{tab:test-std}
\end{table}

\section{Experimental Setup} \label{ssec: Experimental Setup}
\noindent\textbf{Datasets.} 
For image-text experiments, we evaluate our models on four V\&L datasets: VQAv2 \cite{Goyal2017MakingTV} and GQA \cite{DBLP:journals/corr/abs-1902-09506} for visual question answering, NLVR$^{2}$ \cite{DBLP:journals/corr/abs-1811-00491} for visual reasoning, and MSCOCO \cite{DBLP:journals/corr/ChenFLVGDZ15} for image captioning.
As for video-text experiments, we apply our method on four tasks from VALUE \cite{li2021value} benchmark: TVQA \cite{Lei2018TVQALC} and How2QA \cite{Li2020HeroHE} for video question answering, TVC \cite{Lei2020TVRAL} and YC2C \cite{Zhou2018TowardsAL} for video captioning. The statistics of each dataset are shown in \Cref{tab:data}.

\vspace{2pt}
\noindent\textbf{Architecture Details.}
We follow \cite{DBLP:journals/corr/abs-2102-02779} to combine the vision encoder and an encoder-decoder language model to deal with many tasks via the unified text generation framework. For image experiments, we use CLIP-ResNet101 as our vision encoder \cite{DBLP:journals/corr/abs-2103-00020}. Input images are resized to $224 \times 224$ for the memory efficiency. We extract the $7 \times 7$ grid features produced by the last convolutional layer, and then apply adaptive maximum-pooling over the features for downsampling then to $6 \times 6$ for a fair comparison to \cite{DBLP:journals/corr/abs-2102-02779}. For video experiments, we use the features extracted by CLIP (ViT-B/32) following \cite{li2021value}, where they uniformly sample one frame per second and concatenate the CLIP outputs of the sampled frames to form the visual input. We limit the maximum length of visual input to 64 for efficiency. Following \cite{li2021value}, we also include subtitles as additional information.
In addition, we choose BART$_{base}$ \cite{DBLP:journals/corr/abs-1910-13461} as our main encoder-decoder language model, but we also extend the studies to T5$_{base}$ \cite{DBLP:journals/corr/abs-1910-10683}. We use CLIP-BART and CLIP-T5 for representing these two V\&L architectures.

\vspace{2pt}
\noindent\textbf{Training and Evaluation.}
We perform an extensive hyper-parameter search for our models.
See \Cref{sec:hp_search} for details.
We use AdamW to train the model, unless we additionally specify and apply a linear decay scheduler.
We train the models for 20/7 epochs for image-text/video-text tasks,
and warm up the learning rate from 0 to the highest learning rate in the first 2 epochs. In the image-text experiments, we use batch size 500 for CLIP-BART and 250 for CLIP-T5, and the total training time is about 20 hours and 40 hours for CLIP-BART and CLIP-T5 with one A6000 GPU (48G memory), respectively. In video-text experiments, we use batch size 50 for CLIP-BART and the total training time is about 10 hours. We select the last checkpoint for evaluation and report the evaluation score of the four tasks as well as the average score in our experiments. The percentage of updated parameters is also reported as the metric for approaches' efficiency, and we do not take visual encoder into account for computation since it is frozen.

\section{Results and Analysis} \label{sec:results}

\subsection{Multi-Task Parameter-Efficient Fine-Tuning}
Next, we move on to our main experiments on applying parameter-efficient training techniques for V\&L models. We note that the techniques are only used for the language model, as we mentioned in \Cref{ssec:adapters}.
Besides the adapter-based methods and full fine-tuning, we also consider prompt-tuning \cite{Lester2021ThePO} and LoRA \cite{DBLP:journals/corr/abs-2106-09685}, which are competitive parameter-efficient training approaches.
In this case, we also have two variants for each method, that is, \textbf{Single Prompt-Tuning}, \textbf{Multiple Prompt-Tuning}, \textbf{Single LoRA} and \textbf{Multiple LoRA}, where the single one uses the same prompt/low-rank weights for multiple tasks while the multiple one has one prompt/low-rank weights for each task, respectively. The prompts are only added to the input of the encoder because we do not see improvement when the prompts are also added for the decoder. In the following paragraphs, we separately introduce the results in \Cref{tab:main table} and their takeaways. We also report the leaderboard results of multiple approaches on VQA and GQA in \Cref{tab:test-std}, where we select the representative approaches from each family of methods based on their performance in \Cref{tab:main table}. The trend still holds similar to the results in \Cref{tab:main table}.

\vspace{4pt}
\noindent\textbf{Different Visual Representations.}
We compare the results of CLIP-BART and VL-BART (w/o pre-training) and find out that there is a small improvement in using CLIP features over R-CNN features (77.6 vs. 76.7). Note that our CLIP also uses images with a smaller size ($224 \times 224$ vs. $800 \times 1333$), so the result proves the effectiveness of pre-trained cross-modality features.

\vspace{4pt}
\noindent\textbf{Single Adapter Performing the Best.} 
We observe that the vanilla adapter is the most competitive architecture among the three types of adapter variants. We find Half-shared Adapters perform on par with Multiple Adapters with fewer parameters, and the Single Adapter's performance is as competitive as which of the full fine-tuning (77.4 vs. 77.6). The performance boost of Half-shared Adapters and Single Adapter mainly comes from a smaller dataset, NLVR${^{2}}$, and this demonstrates that information sharing benefits the low resource tasks.

We next turn to the results of Hyperformer and Compacter. The Hyperformer shares information across tasks in the hyper-network, thus resulting in that it is more parameter-efficient than the Multiple Adapters (12.22\% vs. 5.79\%).
However, the Single Adapter still outperforms Hyperformer in terms of its parameter-efficiency (4.18\% vs. 5.79\%) and effectiveness (77.4 vs. 76.4). The optimization of hyper-network is harder and it might be one of the reasons to produce this outcome. Compared to Adapter, Compacter does not stand out in our experiments. 

\begin{figure}
    \centering
    \includegraphics[width=\linewidth]{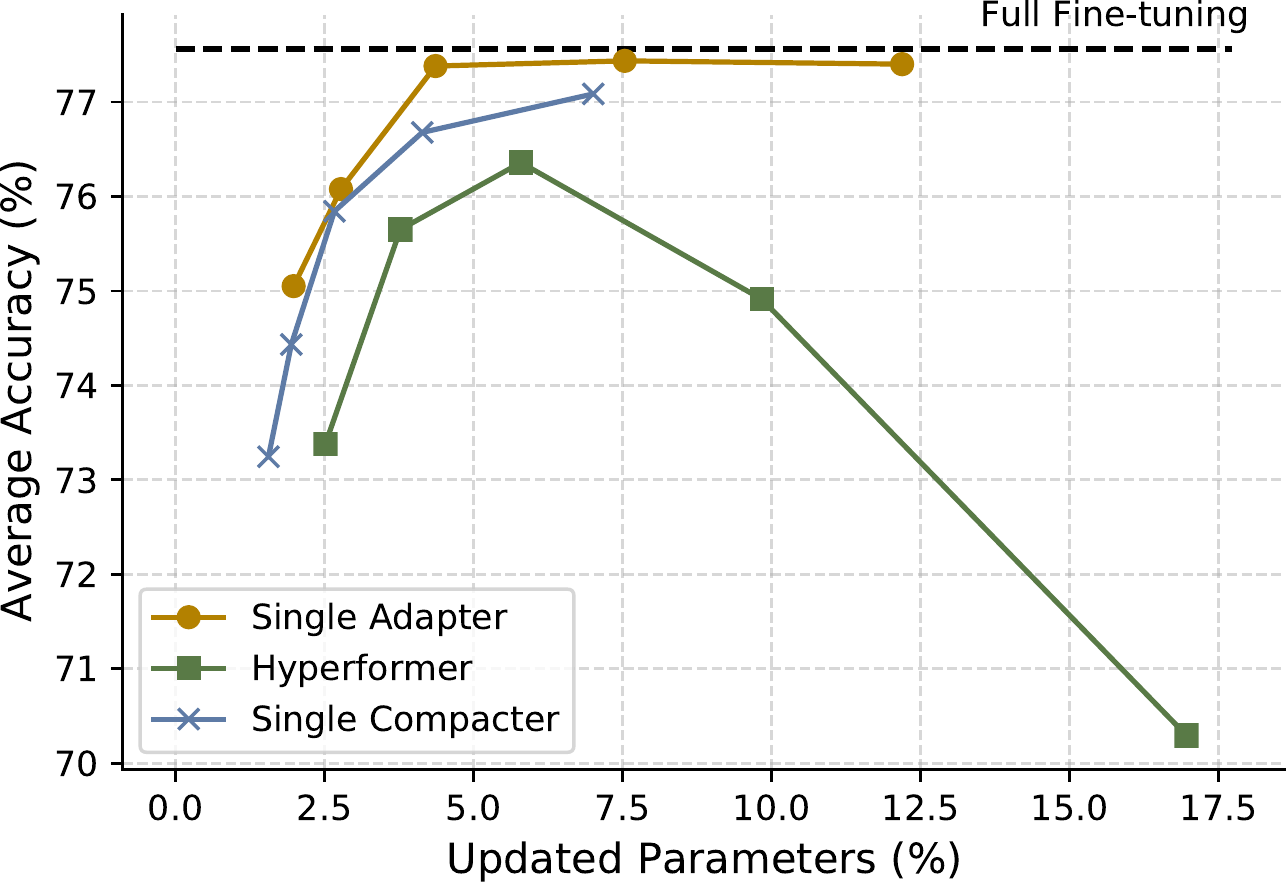}
    \caption{The comparison of three adapter-based approaches over the different percentages of updated parameters. We adjust the hidden dimension $d$ to attain the model with different sizes.}
    \label{fig:adapters_comparison}
\end{figure}

We hypothesize the reason causing the outcomes is: our BART model is pre-trained on pure language tasks, and we would like to adapt the model to perform on V\&L tasks. Nevertheless, the assumption of Kronecker products might be too restrictive, so that Compacter fails to overcome the huge discrepancy between tasks.
To have a complete comparison between three adapter-based approaches, in \Cref{fig:adapters_comparison}, we show their performance over different percentages of updated parameters. We observe that Single Adapter, despite its simple architecture, achieves the best accuracy-efficiency trade-off.

Lastly, we transfer the best configuration of the single adapter to CLIP-T5 and show the results in \Cref{tab:t5}. Note that we use a larger hidden dimension for adapters in this case, and the percentage of updated parameters is 7.98\%. The results conclude that the Single Adapter still achieves a promising accuracy-efficiency trade-off in T5. We leave the results of adding other approaches to T5 in \Cref{sec:hp_t5}.

\vspace{1pt}
\noindent\textbf{LoRA and Prompt-tuning vs. Adapters.}
Compared to the Single Adapter, LoRA uses slightly more parameters and the accuracy drops by approximately 1\%. However, it is still a competitive approach since its performance is better than other methods.
In general, prompt-tuning does not perform well in our experiments. The reason might be similar to Compacter's: because the pre-trained tasks and downstream tasks are dissimilar, the model cannot adapt to the distribution of new datasets with few parameters. Also, prompt-tuning is sensitive to some training configurations such as model size, prompt initialization, and pre-training methods \cite{Lester2021ThePO}. We leave the improvement of prompt-tuning performance for future work.

\noindent\textbf{VL-Adapter in Video-Language Understanding Tasks.}
\Cref{tab:video table} shows the performance of the representative approaches used in \Cref{tab:test-std} on TVQA, How2QA, TVC, and YC2C. We transfer the hyperparameter setup used in \Cref{tab:main table}, but we divide the learning by 3 for all approaches for better performance. Single Adapter again attains the best results among all parameter-efficient methods and is on par with full fine-tuning. From the image-text and video-text experiments, we find that the Adapter is more stable than other approaches.

\begin{table}[t]
    \centering
    \resizebox{\columnwidth}{!}{
    \begin{tabular}{llccc}
    \toprule
    \multirowcell{2}{Type} & \multirowcell{2}{Dataset} & \multicolumn{3}{c}{Data size (\# videos / \# QA pairs, \# captions)} \\
    & & \makecell{Train} & \makecell{Validation} & \makecell{Test}\\
    \midrule
    
    \multirowcell{4}{Image} & VQA \cite{Goyal2017MakingTV} & 113.2K/605.1K & 5.0K/26.7K & 5.0K/26.3K  \\
    & GQA \cite{DBLP:journals/corr/abs-1902-09506}  & 72.1K/943.0K & 10.2K/132.1K & 398/12.6K\\
    & NLVR$^2$ \cite{DBLP:journals/corr/abs-1811-00491} & 103.2K/86.4K & 8.1K/7.0K & 8.1K/7.0K  \\
    & COCO Cap. \cite{DBLP:journals/corr/ChenFLVGDZ15} & 113.2K/566.8K & 5.0K/5.0K & 5.0K/5.0K \\
    
    \midrule

    \multirowcell{4}{Video} & TVQA \cite{Lei2018TVQALC} & 17.4K/122.0K & 2.2K/15.3K &  2.2K/15.3K \\
    & How2QA \cite{Li2020HeroHE} & 24.5K/34.2K & 3.1K/3.1K & 3.1K/3.1K \\
    & TVC \cite{Lei2020TVRAL} & 17.4K/86.7K & 10.8K/43.6K & 10.8K/43.6K \\
    & YC2C \cite{Zhou2018TowardsAL} &  10.3K/10.3K & 3.5K/3.5K & 1.6K/1.6K \\
   
    \bottomrule
    \end{tabular}
    }
    \vspace{-3pt}
    \caption{The statistics of the datasets used in our experiments.}
    \label{tab:data}
    \vspace{-1pt}
\end{table}

\begin{table}[t]
    \centering
    \resizebox{0.9\columnwidth}{!}{
    \begin{tabular}{lccccc}
    \toprule
    \makecell{Method} &  \makecell{VQA} & \makecell{GQA} & \makecell{NLVR$^{2}$} & \makecell{COCO\\Cap.} &  \makecell{Avg.}\\                              
    \midrule
    CLIP-T5 \\
    + Full fine-tuning & 67.3 & \textbf{56.5} & \textbf{75.4} & \textbf{113.1} & \textbf{78.1}\\
    + Single Adapter  & \textbf{67.6} & 56.2 & 73.9 & 111.8 & 77.4 \\
    \bottomrule
    \end{tabular}
    }
    \vspace{-3pt}
    \caption{Results of Single Adapter \& Full fine-tuning on CLIP-T5.}
    \label{tab:t5}
    \vspace{-1pt}
\end{table}

\begin{table}[]
    \centering
    \resizebox{0.85\columnwidth}{!}{
    \begin{tabular}{lcc}
    \toprule
                    & fine-tuning BART & frozen BART \\
    \midrule
        fine-tuning CLIP  &  65.6   & 39.4 \\
        frozen CLIP       &  64.7   & 39.1 \\
    \bottomrule
    \end{tabular}
    }
    \vspace{-2pt}
    \caption{The VQA results of all possible pairs of full fine-tuning or freezing the CLIP and BART. Note that we only train the visual project layer ($\theta^{V \rightarrow L}$) in ``frozen CLIP + frozen BART''.
    }
    \label{tab:training resnet}
\end{table}

\begin{table*}[t]
    \centering
    \resizebox{0.83\textwidth}{!}{
    \begin{tabular}{l c c c c c c}
    \toprule
    \makecell{Method} &
    \makecell{Updated \\ Params \\ (\%)} & \makecell{TVQA \\ test-public \\ Acc. (\%)} & \makecell{How2QA \\ test-public \\ Acc. (\%)} & \makecell{TVC \\ test-public \\ CIDEr} & \makecell{YC2C \\ test-public \\ CIDEr} & \makecell{Avg.} \\
    \midrule
    CLIP-BART \\
    + Full fine-tuning & 100.00 & 76.3 & \textbf{73.9} & 45.7 & \textbf{154.0} & \textbf{87.4} \\
    + Single Adapter      & 3.39 & \textbf{76.6} & \textbf{73.9} & \textbf{46.3} & 152.9 & \textbf{87.4} \\
    + Single LoRA & 5.17 & 75.5 & 72.9 & 44.6 & 140.9 & 83.4 \\
    + Single Prompt & 1.18 & 32.1 & 30.2 & 29.5 & 102.0 & 48.4 \\
    \bottomrule
    \end{tabular}
    }
    \caption{
    The multi-task evaluation results on TVQA, How2QA, TVC and YC2C between full fine-tuning, Single Adapter, LoRA and prompt-tuning.
    }
    \label{tab:video table}
    \vspace{-5pt}
\end{table*}

\subsection{Training or Freezing Visual Encoder} \label{ssec:freezing resnet}
Because the use of CLIP enables the entire model to be end-to-end trainable with stability \cite{DBLP:journals/corr/abs-2107-06383}, we conduct experiments to test the performance of four different training scenarios: all possible pairs of training (full fine-tuning) or freezing CLIP and BART. Every combination is trained and evaluated on VQA. We apply the training tricks\footnote{We have our best jointly training model with a $1 \times 10^{-6}$ learning rate and the SGD optimizer on vision encoder while $1 \times 10^{-4}$ and the AdamW optimizer for the rest of the model.} used in \cite{DBLP:journals/corr/abs-2004-00849, DBLP:journals/corr/abs-2107-06383} and report the results in \Cref{tab:training resnet}.
The results show that there is only 1\% improvement with adding the CLIP into training. Given this result, the advantage of adding adapters inside CLIP over keeping it frozen is limited. Therefore, we have determined to freeze the CLIP consistently to save memory, and this also ideally fits our purpose of training models efficiently. We also report the result of our version of ``Frozen''~\cite{DBLP:journals/corr/abs-2106-13884} \footnote{However, unlike Frozen using both captioning and VQA data, we only use the latter in our experiments.}, where we freeze BART and only fine-tune CLIP. However, the accuracy of 39.4 (in \Cref{tab:training resnet}) is far from that of updating BART, suggesting that fine-tuning the language model is still crucial.

\begin{table}[t]
    \centering
    \resizebox{0.95\columnwidth}{!}{
    \begin{tabular}{lccccc}
    \toprule
    \makecell{Method} &  \makecell{VQA} & \makecell{GQA} & \makecell{NLVR$^{2}$} & \makecell{COCO\\Cap.} &  \makecell{Avg.}\\                              
    \midrule
    $\theta^{V \rightarrow L}$ only    & 32.2 & 25.6 & 52.1 & 78.5 & 47.1 \\
    $\theta^{V \rightarrow L}$ + Layer Norm & 49.5 & 40.1 & 52.4 & 109.6 & 62.9 \\
    Single Adapter                     & 65.9 & 54.5 & 74.2 & 114.9 & 77.4 \\
    \bottomrule
    \end{tabular}
    }
    \vspace{-3pt}
    \caption{Ablation on adding components while adapter training.
    }
    \label{tab:ablation on trainable components}
    \vspace{-1pt}
\end{table}

\subsection{Ablations and Analysis}

In the following paragraphs, we sequentially present our ablation studies and analysis on (1) the contribution of different modules. (2) adapters with task-specific prompts. (3) effect of V\&L  pre-training on adapters. (4) adapters on the discriminative V\&L model.

\vspace{3pt}
\noindent\textbf{Contribution of Modules ($\theta^{V \rightarrow L}$, Layer Norm).}
Recall that our adapter training not only includes the adapters' modules but layer normalization layers and the visual projection layer $\theta^{V \rightarrow L}$. 
To have a better understanding of the contribution to the accuracy of adapters, we conduct ablation studies to gradually add trainable modules and compare their results in \Cref{tab:ablation on trainable components}. 
We observe that only updating $\theta^{V \rightarrow L}$ produces insufficient results, suggesting the need to update the language model partially. We do find a considerable improvement with updating layer normalization layers, but the accuracy is still much behind that of the full fine-tuning, and this result also displays the effectiveness of adapters. We note that this finding deviates from the conclusion in \cite{anonymous2022how}, where updating layer normalization layer is comparable or even better than training adapters inside CLIP for the image classification task.

\vspace{3pt}
\noindent\textbf{Adapters with Task-specific Prompts.}
We experiment to remove task-specific prompts before the input sequence, namely, from ``[task]: [input]'' to ``[input]'', where [task] is task indicator, such as ``vqa'', ``gqa'', ``nlvr'', and ``caption''. The ablation is only for the approaches using one set of parameters for multi-tasking, such as full fine-tuning and Single Adapter. We exclude
Hyperformer in this experiment since we follow the original implementation to remove all prompts and use task embedding as the condition. 
The results of whether to use task-specific prompts are displayed in \Cref{tab:ablation on prompts}. We find that using prompts can improve performance, and the improvement likely comes from resolving the confusion between tasks. However, the model still performs well without prompts. We hypothesize that the data distribution between tasks is large enough for the model to understand to treat them differently, so the added prompts might become redundant. For example, there is no text input in MSCOCO, while there are two input images in NLVR$^2$.

\begin{table}[t]
    \centering
    \resizebox{\columnwidth}{!}{
    \begin{tabular}{lccccc}
    \toprule
    \makecell{Method} &  \makecell{VQA} & \makecell{GQA} & \makecell{NLVR$^{2}$} & \makecell{COCO\\Cap.} &\makecell{Avg.}\\  
                            
    \midrule
    CLIP-BART \\
    - w/o prompt & 66.7 & 56.5 & \textbf{73.2} & 112.4 & 77.2 \\
    - w/ prompt & \textbf{67.6} & \textbf{56.7} & 73.0 & \textbf{112.9} & \textbf{77.6} \\
    \midrule
    CLIP-BART + Single Adapter \\
    - w/o prompt & 65.1 & 53.9 & 72.7 & \textbf{115.6} &	76.8 \\
    - w/ prompt & \textbf{65.9} & \textbf{54.5} & \textbf{74.2} & 114.9 & \textbf{77.4} \\
    \bottomrule
    \end{tabular}
    }
    \vspace{-3pt}
    \caption{Ablation results of adding task-specific prompts.}
    \label{tab:ablation on prompts}
\end{table}

\begin{table}[t]
    \centering
    \resizebox{0.85\columnwidth}{!}{
    \begin{tabular}{lccccc}
    \toprule
    \makecell{Method} &  \makecell{VQA} & \makecell{GQA} & \makecell{NLVR$^{2}$} & \makecell{COCO\\Cap.} &\makecell{Avg.}\\ 
    \midrule
    VL-BART \cite{DBLP:journals/corr/abs-2102-02779} & 69.1 & \textbf{59.0} &	73.3 & 111.5 & 78.2 \\
    CLIP-BART \\
    + Full fine-tuning   & 69.2 & 57.5 & \textbf{75.0} & 112.1 & 78.5 \\
    + Single Adapter   & \textbf{69.4} & 58.1 & 73.7 & \textbf{115.7} & \textbf{79.2} \\
    \bottomrule
    \end{tabular}
    }
    \vspace{-2pt}
    \caption{The fine-tuning results of full fine-tuning and Single Adapter after pre-training on V\&L tasks first.}
    \label{tab:ablation on pretraining}
    \vspace{-5pt}
\end{table}

\vspace{3pt}
\noindent\textbf{Effect of V\&L Pre-training on Adapters.}
Since this work is about low-cost training, most of our experiments are based on the weights without V\&L pre-training.
This results in a performance gap between the approaches in this paper and \textit{state-of-the-art} ones. We therefore also explore whether adapter training can take advantage of V\&L pre-trained weights. We follow \cite{DBLP:journals/corr/abs-2107-06383} to pre-train on COCO \cite{lin2014microsoft} and VG \cite{Krishna2016VisualGC} images with multi-modal language modeling, visual question answering, and image-text matching. We exclude the referring tasks (grounded captioning and visual grounding) because they need bounding box information, which cannot be obtained using CLIP. Refer to the training details in \cite{DBLP:journals/corr/abs-2107-06383}.
\Cref{tab:ablation on pretraining} shows the fine-tuning results with V\&L pre-training. In this case, the Single Adapter even is more competitive than full fine-tuning with only a few parameters being trained, suggesting that the adapters work well with different kinds of pre-trained weights.

\vspace{3pt}
\noindent\textbf{Adapters on the Discriminative V\&L Pre-trained Model.}
While most of our experiments and ablations are conducted on generative V\&L models, we also applied adapters to CLIP-ViL \cite{DBLP:journals/corr/abs-2107-06383}, a SOTA discriminative model, and report the results of VQA \cite{Goyal2017MakingTV}, SNLI-VE \cite{Xie2019VisualEA}, and GQA \cite{DBLP:journals/corr/abs-1902-09506} in \Cref{tab:clip-vil}. Adapters are also not added into CLIP due to the same reason in \Cref{ssec:freezing resnet}. Note that Single Adapter only updates from 4.3 to 6.2\% of parameters with only a small gap in accuracy compared to full fine-tuning, which demonstrates the effectiveness of adapters for discriminative architectures.

\begin{table}[t]
    \centering
    \resizebox{0.85\columnwidth}{!}{
    \begin{tabular}{l c c c c}
    \toprule
    \makecell{Method} &
    \makecell{Updated \\ Params \\ (\%)} &
    \makecell{VQA \\ test-dev} & \makecell{SNLI-VE \\ dev} & \makecell{GQA \\ test-dev}  \\
                            
    \midrule
    Full Fine-tuning     & 100.0 & 76.9 & 80.7 & 61.8  \\
    Single Adapter  & 4.3 - 6.2 & 76.5 & 80.2 & 60.7 \\
    \bottomrule
    \end{tabular}
    }
    \vspace{-3pt}
    \caption{The results of adding adapters to CLIP-ViL on VQA, SNLI-VE, and GQA. Note that the number of parameters vary across tasks due to different output heads.}
    \label{tab:clip-vil}
\end{table}

\section{Discussion and Conclusion}
We conduct comprehensive studies on the evaluation of three adapter-based approaches on challenging V\&L (image-text and video-text) tasks. We employ a unified format and architecture to solve the tasks in a multi-tasking learning setup. With a thorough hyper-parameter search, we benchmark the performance of those methods and find the Single Adapter, which is a shared-weight vanilla adapter, is the best in terms of accuracy, efficiency, and simplicity. We have also shown that Single Adapter works well with a state-of-the-art discriminative model (CLIP-ViL). Lastly, we conduct ablation studies on understanding the contribution of different trainable modules, adapters with task-specific prompts, and the effect of V\&L pre-training on adapters.

Next, we discuss some limitations of this work. We have carried out extensive experiments on four V\&L tasks with our proposed CLIP-BART and CLIP-T5. However, different architectures have their own best hyper-parameters, and data distributions are varied across tasks, so our results and findings do not always guarantee to be generalized to other tasks. Furthermore, we experiment with the three popular adapter variants, but they cannot represent all the adapter-based approaches.

\section*{Acknowledgments}
We thank the reviewers, Hyounghun Kim, Gedas Bertasius, and Hao Tan for their helpful discussions.
This work was supported by ARO  Award W911NF2110220, and ONR Grant N000141812871, and Google Focused Research Award. The views, opinions, and/or findings contained in this article are those of the authors and not of the funding agency.

{\small
\bibliographystyle{ieee_fullname}
\bibliography{egbib}
}

\vspace{0.5cm}
\appendix

\begin{table*}[t]
    \centering
    \resizebox{0.9\textwidth}{!}{
    \begin{tabular}{l c c c c c c c}
    \toprule
    \makecell{Method} &
    \makecell{Best \\ Learning Rate} &
    \makecell{Updated \\ Params \\ (\%)} & \makecell{VQA \\ Karpathy test \\ Acc. (\%)} & \makecell{GQA \\ test-dev \\ Acc. (\%)} & \makecell{NLVR$^{2}$ \\ test-P \\ Acc. (\%)} & \makecell{COCO Cap. \\ Karpathy test \\ CIDEr} & \makecell{Avg.} \\
    \midrule
    (A) Full fine-tuning & $1 \times 10^{-4}$ & 100.00 & 67.6 & 56.7 & 73.0 & 112.9 & \textbf{77.6} \\
    \midrule
    (B) Multiple Adapters \\
    (B.1) - $\bm{d = 96}$  & $3 \times 10^{-4}$ & 12.22 & 65.4 & 54.0 & 69.8 & 114.3	& \textbf{75.9} \\
    (B.2) - $d = 48$  & $1 \times 10^{-3}$ & 7.58 & 65.4 & 53.7 & 65.3 & 115.0  & 74.9\\
    \midrule
    (C) Half-shared Adapters ($d = 96$) \\
    (C.1) - sharing downsampling layers & $3 \times 10^{-4}$ & 8.40 & 65.2 & 53.3 & 70.2 & 113.8 & 75.6 \\
    (C.2) - \textbf{sharing upsampling layers} & $3 \times 10^{-4}$ & 8.36 & 65.2 & 53.4 & 71.2 & 113.7 & \textbf{75.9} \\
    \midrule
    (D) Single Adapter \\   
    (D.1) - $d = 192$ & $1 \times 10^{-3}$ & 7.54 & 66.5 & 54.0 & 73.5 & 115.8 & 77.4 \\
    (D.2) - $\bm{d = 96}$ & $1 \times 10^{-3}$ & 4.36 & 65.9 & 54.5 & 74.2 & 114.9 & \textbf{77.4} \\
    (D.3) - $d = 64$ & $1 \times 10^{-3}$ & 3.30 & 65.2 & 53.8 & 72.3 & 114.5 & 76.4 \\
    (D.4) - $d = 48$ & $1 \times 10^{-3}$ & 2.78 & 64.7 & 53.9 & 71.5 & 114.2 & 76.1 \\
    (D.5) - $d = 24$ & $1 \times 10^{-3}$ & 1.98 & 63.5 & 52.2 & 71.0 & 113.5 & 75.1 \\
    \midrule
    (F) Hyperformer \\
    (F.1) - $\bm{d = 96, d_p = 8}$ & $1 \times 10^{-3}$ & 5.79 & 65.1 & 53.4 & 72.3 & 114.6 & \textbf{76.4} \\
    (F.2) - $d = 96, d_p = 4$ $^{*}$ & $1 \times 10^{-3}$ & 3.87 & 65.0 & 53.2 & 51.1 & 114.9 & 71.0 \\
    (F.3) - $d = 48, d_p = 8$ & $1 \times 10^{-3}$ & 3.77 & 64.5 & 52.5 & 71.3 & 114.3 & 75.7 \\
    \midrule
    (G) Multiple Compacters ($d = 48$)  \\
    
    (G.1) - w/ sharing weights, w/ low-rank param. ($r = 1$), $k = 1$ & $1 \times 10^{-3}$ & 1.381 & 50.8 & 41.6 & 53.5 & 104.9 & 62.7 \\
    (G.2) - w/ sharing weights, w/ low-rank param. ($r = 1$), $k = 4$ & $1 \times 10^{-3}$ & 1.381 & 52.6 & 43.5 & 54.0 & 111.6 & 65.4 \\
    (G.3) - w/ sharing weights, w/ low-rank param. ($r = 1$), $k = 8$	& $1 \times 10^{-3}$ & 1.382 & 52.2 & 42.4 & 58.3 & 109.8 & 65.7 \\
    (G.4) - w/ sharing weights, w/ low-rank param. ($r = 1$), $k = 12$ & $1 \times 10^{-3}$ & 1.383 & 53.9 & 43.7 & 60.4 & 111.1 & 67.3 \\
    
    (G.5) - w/ sharing weights, w/o low-rank param., $k = 4$ & $1 \times 10^{-3}$ & 2.83 & 52.7 & 42.7 & 59.7 & 112.2 & 66.8 \\
    
    (G.6) - w/o sharing weights, w/o low-rank param., $k = 2$ & $1 \times 10^{-3}$ & 4.42 & 64.0 & 52.9 & 68.3 & 115.7 & 75.2 \\
    (G.7) - w/o sharing weights, w/o low-rank param., $k = 4$ & $1 \times 10^{-3}$ & 2.84 & 62.4 & 51.4 & 68.6 & 115.5 & 74.5 \\
    (G.8) - w/o sharing weights, w/o low-rank param., $k = 8$ & $1 \times 10^{-3}$ & 2.11 & 61.4 & 50.9 & 68.9 & 115.4 & 74.1 \\
    
    (H) Multiple Compacters ($d = 96$)  \\
    (H.1) - \textbf{w/o sharing weights, w/o low-rank param., }\bm{$k = 2$} & $1 \times 10^{-3}$ & 7.02 & 64.6 & 53.4 & 69.1 & 116.0 & \textbf{75.8} \\
    \midrule
    
    (I) Single Compacter ($d = 96$)  \\
    (I.1) - \textbf{w/o sharing weights, w/o low-rank param., }\bm{$k=2$} & $1 \times 10^{-3}$ & 2.67 & 64.2 & 53.3 & 71.7 & 114.1 & \textbf{75.8} \\
    
    Single Compacter ($d = 48$)  \\
    (I.2) - w/o sharing weights, w/o low-rank param., $k=2$ & $1 \times 10^{-3}$ & 1.59 & 61.6 & 50.7 & 69.0 & 114.0 & 73.8 \\
    \midrule
    (J) Multiple Prompts \\
    (J.1) - \bm{$N_p = 40, d_m = 800$} & $1 \times 10^{-3}$ & 4.53 & 43.8 & 38.1 & 51.1 & 104.6 & 59.4 \\
    (J.2) - $N_p = 40, d_m = 100$ & $1 \times 10^{-3}$ & 1.64 & 47.4 & 37.0 & 49.8 & 108.6 & \textbf{60.7} \\
    \midrule
    (K) Single Prompt \\
    
    (K.1) - \bm{$N_p = 40, d_m = 800$} & $1 \times 10^{-3}$ & 2.00 & 44.0 & 36.3 & 51.8 & 103.9 & \textbf{59.0} \\
    (K.2) - $N_p = 40, d_m = 100$ & $1 \times 10^{-3}$ & 1.25 & 43.5 & 36.4 & 52.0 & 103.4 & 58.8 \\
    
    \bottomrule
    \end{tabular}
    }
    \caption{
    The multi-task evaluation results for CLIP-BART on VQA, GQA, NLVR$^{2}$ and COCO Caption between adapter-based approaches with different hyper-parameters. We bold the highest average accuracy separately for each approach, and we also bold the best configuration we used in the main paper. Note that we don't use V\&L pre-training for every model. * denotes the NLVR results might be improved if we use different learning rates.
    }
    \label{tab:bart hyper-parameter}
\end{table*}

In this appendix,
we first explain our prompt-tuning experiment setup with details (\Cref{sec:detail prompt-tuning}).
Then we show the experimental results of the complete hyper-parameter search for the adapter-based and prompt-based techniques (\Cref{sec:hp_search}).

\section{Details for Prompt-tuning}
\label{sec:detail prompt-tuning}

Prompt-tuning \cite{Lester2021ThePO} adds trainable parameters to the encoder's inputs for adapting those parameters for new tasks without changing the model. Specifically, we assume the input indices for generating prompts are $1, 2, ..., N_p \in \mathbb{N}$,
where $N_p$ is the length of prompts.
We next apply a three-layer neural network to transform the prompts embeddings to the correct dimension for the language model. The first layer is an embedding layer, parameterized by $\theta_E \in \mathbb{R}^{N_p \times d_i}$, and the rest of the two layers are parameterized by $\theta_D \in \mathbb{R}^{d_i \times d} $ and $\theta_U \in \mathbb{R}^{d \times d_i}$. Since the architecture of the prompt network is quite similar to the adapter module, we use the same notations as we used in adapters for simplicity. The mathematic form can be written as the following,
\begin{align}
    \begin{split}
    h & = f_{\theta_E}(p) \\
    h_p & = f_{\theta^U}(\sigma( f_{\theta^D}(h)))
    \end{split}
\end{align}
where $p \in {1, 2, ..., N_p}$, $h_p$ being the prompt of index $p$, and we use Tanh as the activation function. 
Next, we can combine the prompt embeddings with vision and sentence embedding, feed-forwarding to the model, and train them with backpropagation. The trainable parameters consist of the input prompts embeddings and the parameters of the three-layer neural network. Note that unlike in adapter modules that $d$ is smaller than $d_i$ for saving memory, $d$ in the prompt network is sometimes greater than $d_i$ since it is the main hyper-parameter to increase the number of trainable parameters. The length of the prompt $N_p$ does not contribute much to the number of parameters since it only influences the embedding layer, which usually is a small layer. Thus, using longer prompts is a parameter-efficient method to train models. However, the memory usage would increase significantly with longer prompts due to the quadratic cost of attention layer on input lengths.
For a fair comparison, we maximize $N_p$ to use the same amount of memory as being used in adapter-based approaches (around 40 GB).

\section{Details for LoRA}

Assume the initial weight for a layer is $\theta^{d_i \times d_o}$, LoRA \cite{DBLP:journals/corr/abs-2106-09685} learns two low-rank matrices $A \in R^{d_i \times d}$ and $B \in R^{d \times d_o}$ ($d \ll d_i, d_o$) to approximate the weight's updates, that is

\begin{equation}
    \Delta \theta = AB
\end{equation}

The output of this layer can be written as $f_{\theta+AB}(h)$.
Hu et al. \cite{DBLP:journals/corr/abs-2106-09685} apply this trick to attention layers (not in feed-forward layers), and they also update bias terms of the model. Compared to the original model, using adapters or prompt-tuning, which modify the network or inputs, causes extra computation in the inference. However, there is no extra overhead in LoRA since we can add the updates back to the model after training.

\section{Hyper-parameter Search}
\label{sec:hp_search}

We search over the learning rates among $\{1 \times 10^{-4}, 3 \times 10^{-4}, 1 \times 10^{-3}\}$ for each hyper-parameter configuration. To reduce the cost of searching, we utilize a heuristic logic: we first search for the best learning rate for the one hyper-parameter configuration (randomly chosen) and then use the same learning rate for other configurations. We perform another learning rate search only if the results are diverged for some tasks (e.g. sometimes the results of NLVR$^2$ become very low at certain learning rates).

For the Adapter, the only hyper-parameter is the hidden dimension $d$. We also ablate two variants of Half-shared Adapters: sharing upsampling or downsampling layers. We include the search about the projected hidden dimension $d_e$ for the task projector network in the Hyperformer. Regarding the Compacter, we have tried different numbers of Kronecker products ($k$), hidden dimension $d$, and whether sharing weights and using low-rank parameterization. We also tune the $d_m$ for prompt-tuning.

\subsection{CLIP-BART Hyper-parameter Search}
\label{sec:hp_bart}

We show the results of the hyper-parameter search in \Cref{tab:bart hyper-parameter}. We bold the final configurations used in the main paper and we also list the configurations in \Cref{tab:final configuration}. The exception is that we use the same hyper-parameters for the ``Single'' and ``Multiple'' approaches. For example, even though Multiple Prompts perform better when $d_m = 100$, we still use $d_m = 800$ for both Multiple Prompts and Single Prompt for consistency (J and K rows in \Cref{tab:bart hyper-parameter}).

\begin{table*}[t]
    \centering
    \resizebox{0.9\textwidth}{!}{
    \begin{tabular}{l c c c c c c c}
    \toprule
    \makecell{Method} &
    \makecell{Best \\ Learning Rate} &
    \makecell{Updated \\ Params \\ (\%)} & \makecell{VQA \\ Karpathy test \\ Acc. (\%)} & \makecell{GQA \\ test-dev \\ Acc. (\%)} & \makecell{NLVR$^{2}$ \\ test-P \\ Acc. (\%)} & \makecell{COCO Cap. \\ Karpathy test \\ CIDEr} & \makecell{Avg.} \\
    \midrule
    (A) Full fine-tuning & $1 \times 10^{-4}$ & 100.00 & 67.3 & 56.5 & 75.4 & 113.1 & \textbf{78.1} \\
    \midrule
    (B) Multiple Adapters \\
    (B.1) - \bm{$d = 192$}  & $1 \times 10^{-3}$ & 24.56 & 66.0 & 55.7 & 51.8 & 111.9 & 71.3 \\
    (B.2) - $d = 96$  & $1 \times 10^{-3}$ & 14.29 & 66.1 & 55.7 & 52.5 & 112.8 & \textbf{71.8} \\
    \midrule
    (C) Single Adapter \\   
    (C.1) - $d = 384$ & $3 \times 10^{-4}$ & 14.25 & 67.6 & 55.9 & 73.6 & 111.8 & 77.2 \\
    (C.2) - \bm{$d = 192$} & $3 \times 10^{-4}$ & 7.98 & 67.6 & 56.2 & 73.9 & 111.8	& \textbf{77.4} \\
    (C.3) - $d = 96$ & $1 \times 10^{-3}$ & 4.49 & 66.4 & 55.5 & 72.7 & 111.5 & 76.5 \\
    (C.4) - $d = 48$ & $1 \times 10^{-3}$ & 2.64 & 65.7 & 54.7 & 70.9 & 111.1 & 75.6 \\
    \midrule
    (D) Hyperformer \\
    (D.1) - \bm{$d = 192, d_p = 8$} & $1 \times 10^{-3}$ & 6.37 & 65.5 & 55.1 & 71.5 & 112.2 & \textbf{76.1} \\
    (D.2) - $d = 192, d_p = 4$& $1 \times 10^{-3}$ & 3.99 & 65.0 & 53.9 &	 70.4 & 111.7 & 75.2 \\
    \midrule
    (E) Multiple Compacters ($d = 192$)  \\
    (E.1) - \textbf{w/o sharing weights, w/o low-rank param., }\bm{$k = 2$} $^{*}$ & $1 \times 10^{-3}$ & 14.30 & 66.1 & 55.0 & 52.1 & 112.9 & 71.5 \\
    (E.2) - w/o sharing weights, w/o low-rank param., $k = 4$ $^{*}$ & $1 \times 10^{-3}$ & 8.06 & 65.4 & 55.0 & 52.2 & 113.2 & \textbf{71.5} \\
    (E.3) - w/o sharing weights, w/o low-rank param., $k = 8$ $^{*}$ & $1 \times 10^{-3}$ & 4.66 & 63.3 & 52.9 & 51.7 & 110.4 & 69.6 \\
    \midrule
    
    (F) Single Compacter ($d = 192$)  \\
    (F.1) - \textbf{w/o sharing weights, w/o low-rank param., }\bm{$k=2$} & $1 \times 10^{-3}$ & 4.49 & 67.0 & 56.6 & 72.5 & 112.7 & \textbf{77.2} \\
    (F.2) - w/o sharing weights, w/o low-rank param., $k=4$ & $1 \times 10^{-3}$ & 2.65 & 66.1 & 55.2 & 71.8 & 111.7 & 76.2 \\
    (F.3) - w/o sharing weights, w/o low-rank param., $k=8$ & $1 \times 10^{-3}$ & 1.72 & 65.2 & 54.1 & 71.6 & 111.5 & 75.6 \\

    \bottomrule
    \end{tabular}
    }
    \caption{
    The multi-task evaluation results for CLIP-T5 on VQA, GQA, NLVR$^{2}$ and COCO Caption between adapter-based approaches with different hyper-parameters. We bold the highest average accuracy separately for each approach, and we also bold the best configuration we used in the main paper. Note that we don't use V\&L pre-training for every model. * denotes the NLVR$^{2}$ results might be improved if we use different learning rates.
    }
    \label{tab:t5 hyper-parameter}
\end{table*}

\begin{table*}[t]
    \centering
    \resizebox{0.9\textwidth}{!}{
    \begin{tabular}{c l c c c}
    \toprule
    \makecell{Model} & \makecell{Approach} & \makecell{Learning Rate} & \makecell{Batch size} & \makecell{Other hyper-parameters} \\
    \midrule
    
    \multirow{9}{*}{CLIP-BART} & Full fine-tuning & $1 \times 10^{-4}$ & 500 & - \\
    
    & Multiple Adapters & $3 \times 10^{-4}$ & 500 & $d = 96$\\
    & Half-shared Adapters & $3 \times 10^{-4}$ & 500 & sharing upsampling layers, $d = 96$ \\ 
    & Single Adapter & $1 \times 10^{-3}$ & 500 & $d = 96$ \\
    & Hyperformer & $1 \times 10^{-3}$ & 500 & $d = 96, d_p = 8$ \\
    & Multiple Compacters & $1 \times 10^{-3}$ & 500 & remove share weight and low-rank, $d = 96, k = 2$ \\
    & Single Compacter & $1 \times 10^{-3}$ & 500 & remove share weight and low-rank, $d = 96, k = 2$ \\
    & Multiple Prompts & $1 \times 10^{-3}$ & 500 &  $N_p = 40, d_m = 800$ \\
    & Single Prompt & $1 \times 10^{-3}$ & 500 &  $N_p = 40, d_m = 800$ \\

    \midrule
    
    \multirow{6}{*}{CLIP-T5} & Full fine-tuning & $1 \times 10^{-4}$ & 250 & - \\
    
    & Multiple Adapters & $1 \times 10^{-3}$ & 250 & $d = 192$\\
    
    & Single Adapter & $3 \times 10^{-4}$ & 250 & $d = 192$ \\
    & Hyperformer & $1 \times 10^{-3}$ & 250 & $d = 192, d_p = 8$ \\
    & Multiple Compacters & $1 \times 10^{-3}$ & 250 & remove share weight and low-rank, $d = 192, k = 2$ \\
    & Single Compacter & $1 \times 10^{-3}$ & 250 & remove share weight and low-rank, $d = 192, k = 2$ \\
    
    \bottomrule
    \end{tabular}
    }
    \caption{
    The best hyperparameter configurations for different parameter-efficient training approaches.
    }
    \label{tab:final configuration}
\end{table*}

\subsection{CLIP-T5 Hyper-parameter Search}
\label{sec:hp_t5}
We display the results of the hyper-parameter search for CLIP-T5 in \Cref{tab:t5 hyper-parameter} and final configurations for each method in \Cref{tab:final configuration}. We find that the Compacter (F.1 in \Cref{tab:t5 hyper-parameter}) shows the different fashion in T5: it can perform similarly to the Single Adapter (C.2 in \Cref{tab:t5 hyper-parameter}) using fewer parameters. This might because the Compacter is mainly validated on T5 in \cite{DBLP:journals/corr/abs-2106-04647}.

\section{Learderboard Results of the Test-dev split for VQA} \label{sec:test-dev vqa}

While we use the Karpathy split for VQA evaluation in the main content, we also report the test-dev results for all approaches in \Cref{tab:test-dev vqa}. In short, the trend remains similar as using the Karpathy split, and the Single Adapter still performs the best among parameter-efficient training methods.

\begin{table*}[t]
    \centering
    \resizebox{\textwidth}{!}{
    \begin{tabular}{l c c c c c c c c c c c}
    \toprule
    \makecell{Datasets} &
    \makecell{Full \\ fine-tuning} & \makecell{Multiple \\Adapters} & \makecell{Half-shared \\Adapters} & \makecell{Single \\ Adapter} & \makecell{Hyperformer} & \makecell{Multiple \\Compacters} & \makecell{Single \\Compacter} & \makecell{Multiple \\LoRA} & \makecell{Single \\LoRA} & \makecell{Multiple \\Prompts} & \makecell{Single \\Prompt}\\
    \midrule
    VQA & 69.9 & 66.7 & 66.6 & 68.1 & 67.5 & 66.5 & 66.4 & 67.4 & 67.0 & 48.8 & 45.4\\
    \bottomrule
    \end{tabular}
    }
    \caption{
    Test-dev results for VQA.
    }
    \label{tab:test-dev vqa}
    \vspace{-5pt}
\end{table*}

\end{document}